\documentclass{article}
\usepackage{ijcai19}

\usepackage{times}
\usepackage{soul}
\usepackage{url}
\usepackage[hidelinks,draft]{hyperref}
\usepackage[utf8]{inputenc}
\usepackage[small]{caption}
\usepackage{graphicx}
\usepackage{algorithm}
\usepackage{algorithmic}
\usepackage{multirow}
\usepackage{latexsym}
\usepackage{tabularx}
\usepackage{enumitem}
\usepackage{amsmath, amsfonts, amssymb}
\usepackage{capt-of}
\usepackage{array, booktabs}
\usepackage{subcaption}
\usepackage{float}

\usepackage{macros}
\usepackage{mydefs}

\begin{document}

\title{Estimating Causal Effects of Tone in Online Debates}
\author{
	Dhanya Sridhar$^1$
	\and
	Lise Getoor$^2$
	\affiliations
	$^1$Columbia University\\
	$^2$UC Santa Cruz\\
	\emails
	ds3778@columbia.edu
}

\maketitle
\begin{abstract}
Statistical methods applied to social media posts shed light on the dynamics of online dialogue.
For example, users' wording choices predict their persuasiveness \cite{tan2016winning,danescu2012you} and users adopt the language patterns of other dialogue participants \cite{danescu2011mark,danescu2012echoes}.
In this paper, we estimate the causal effect of reply tones in debates on linguistic and sentiment changes in subsequent responses.
The challenge for this estimation is that a reply's tone and subsequent responses are confounded by the users' ideologies on the debate topic and their emotions.
To overcome this challenge, we learn representations of ideology using generative models of text.
We study debates from \Forumcom~and compare annotated tones of replying such as emotional versus factual, or reasonable versus attacking.
We show that our latent confounder representation reduces bias in ATE estimation.
Our results suggest that factual and asserting tones affect dialogue
and provide a methodology for estimating causal effects from text. 

\end{abstract}
\section{Introduction}
\label{sec:intro}

Debates on online forums or social media sites provide observational data for studying discourse.
Current understanding draws upon theories such as linguistic accommodation, which states that dialogue participants change and vary their wording styles to mirror one another \cite{gallois2015communication,giles2008communication}.
Statistical methods applied to social media have shown evidence of linguistic style accommodation \cite{danescu2011mark}, power dynamics \cite{danescu2012echoes} and varying persuasiveness of argumentation styles \cite{tan2016winning}.

In this paper, we focus on online debates.
We ask the causal question of how the tone used to reply in a debate affects changes in linguistic style and sentiment.
To illustrate the setting, consider a snippet of a debate between two users, A and B, on a given topic. User A posts her opinion on the topic to which user B replies
with a nasty tone. User A writes a second post, responding to B's post. The goal is to examine the change in A's sentiment or linguistic style between her first and second post.
For example, we may observe that between her first post and her second post in response to B, A's negative sentiment increased.
We study how A's sentiment might have changed had B been nice instead of nasty in the reply.
We consider such sequences of three posts within debates and cast the tone of the first reply as the treatment.
Formally, we estimate the average treatment effect of reply tone on changes in sentiment and linguistic style.

The challenge for this estimation is that the ideologies encoded in A and B's posts, and A's initial sentiment affect both B's reply tone and A's subsequent response.
For example, consider a debate between A and B on gun control. 
Examples of opposing ideologies that influence the debate are strong opposition to gun violence versus strict interpretations of the constitution.
Ideological differences or innate negativity from A provoke both B's nasty tone and A's subsequent reactions.
Valid causal inference requires adjusting for these confounders when estimating the treatment effect. 
While sentiment analysis tools are available for extracting posts' sentiment, modeling the latent ideologies that underpin a particular debate requires careful consideration.
This paper proposes representations of ideologies learned from debates to adjust for confounding.

In recent social media analyses, adjusting for attributes such as discussion topic, 
post authors, timing of posts and posting frequency has been useful to understand post likeability, antisocial behavior and emoji use \cite{tan2014effect,jaech7talking,cheng2015antisocial,pavalanathan2015emoticons}.
The adjustments are typically performed by only comparing posts that have similar values of the confounder.
Our approach requires adjusting for the underlying facets of a debate, an unobserved and multi-dimensional confounder.
We use a generative model of text to learn latent representations of posts to this end. 

\paragraph{Main Idea.} 
The goal of this paper is to estimate the causal effects of tones used in debate replies on other users' change in linguistic style and sentiment.
We identify treatments (tone) and outcome representations which capture the change in sentiment and linguistic style across sequences of posts.
We use three plug-in estimators of average treatment effects: regression, inverse propensity weighting (IPW), and augmented IPW.
To adjust for confounding in these estimates, we find latent representations of posts that capture the underlying ideological viewpoints of the debate.
Our contributions include:
\begin{itemize}
	\item Formulating the problem of estimating the effects of tone on subsequent dialogue within the framework of causal inference.
	\item Learning latent representations of ideologies in debates from generative language models to represent confounders.
	\item Validating the consistency of estimated effects using three different estimators and examining multiple tones of reply in online debates.
\end{itemize}
We study \Forumcom, an online debate forum corpus that includes annotations for multiple reply tones including nasty versus nice, or emotional versus factual \cite{Walkeretal12c}.
Through comparisons against a naive confounder representation and studies across reply tones, we examine the implications of various modeling choices.
With these findings, we highlight guidelines for estimating treatment effects using text from social media.
We also find that factual replies significantly affect how users' vary their linguistic style and sentiment between posts.
\section{Related Work}
\label{sec:related}

Prior work on online debate forums primarily focus on supervised prediction tasks.
Debate text and reply structure between users has been used to predict stance, sentiment and reply polarity \cite{Abbottetal11,Walkeretal12a,hasan2013extra,sridhar2015joint,MisraWalker13,rosenthal2015couldn}.
Related work has also used the Change My View forum on Reddit.com to predict persuasiveness from styles of argumentation and characterize
logical fallacies \cite{tan2016winning,wei2016post,habernal2018before}.

Similarly, unsupervised methods have been applied to analyze dialogue.
Statistical models have been proposed to quantify linguistic accommodation both on Twitter and  U.S. Supreme Court arguments \cite{danescu2011mark,danescu2012echoes}. 
In contrast, we formulate an approach based on causal inference.
 
Existing work on applying causal inference methods to social media focuses on controlling for confounding, or inferring treatments and outcomes from text.
One line of work controls for observable confounders such as topic \cite{tan2014effect,jaech7talking}, timing of posts \cite{jaech7talking} and the post author \cite{tan2014effect}. 
Another line of research uses social media posts to estimate the effect of exercise on mood by inferring both exercise habits and mood from text \cite{dos2015using,olteanu2017distilling}.
In a different line of work, embeddings of text have been used as proxy confounders to study causal effects on paper acceptance \cite{veitch2019using}.

\section{Technical Background}

We review estimating the average treatment effect (ATE) for binary treatments 
from observational data.
We have $n$ iid observations called units, $i = 1 \ldots n$.
Each unit is treated or not, and we denote this treatment assigment $T_i \in \{0,1\}$.
We say $Y_i(1)$ is the potential outcome if we treat unit $i$ (set $T_i=1$),
and analogously, $Y_i(0)$ if we do not treat $i$ (set $T_i=0$).
The average treatment effect (ATE) compares potential outcomes:
\begin{equation}
ATE = \mathbb{E}[Y(1)] - \mathbb{E}[Y(0)]
\end{equation}
However, we only observe one outcome for each unit, conditioned on its assigned treatment $Y_i(T_i) | T_i$.
If we compute the ATE above by simply averaging over treated and untreated populations, 
the estimate will typically be biased because $Y(0), Y(1)$ are not independent of the assigned treatments $T$.
Put simply, knowing the treated and untreated units gives us information about their outcomes.

In observational studies, this bias occurs because variables $Z$, called confounders, may affect both the treatment and outcome.
If we observe the confounders for each unit, $Z_i$, then we have $Y(0), Y(1) \perp T | Z$, the condition called ignorability.
In this case, the ATE, which we denote $\psi$, is identifiable as a parameter of the observational distribution
by a theorem called adjustment:
\begin{equation}
\psi = \mathbb{E}_{Z} \big[\mathbb{E}_{Y|T=1}[Y|Z, T=1] - \mathbb{E}_{Y|T=0}[Y|Z, T=0]]
\end{equation}
In plain English, the ATE is: 
how do treated and control units differ in outcome when we average over the varying rates at which units receive treatment?
We refer to work by Pearl and Rubin for an in-depth treatment of causal inference \cite{pearl2009causality,rubin2005causal}.

\subsection{Estimators for ATE}
Drawing from extensive work on estimating ATEs, we present three estimators for $\psi$. We will return to using these
estimators in our empirical study.
The first estimator fits expected outcomes $Q(Z, T) = \mathbb{E}[Y|Z, T]$ from the observations, e.g., with linear regression. 
The corresponding ATE estimate,
$\mle$ is:
\begin{equation}
\mle = \frac{1}{n} \sum_{i=1}^{n} Q(Z_i, 1) - Q(Z_i, 0)
\end{equation}

The second estimator reweights observed outcomes using the propensity score, $P(T=1|Z)$. The resulting inverse propensity weighting (IPW) estimator is:
\begin{equation}
\ipw = \frac{1}{n} \sum_{i=1}^{n} \frac{Y_iT_i}{P(T_i=1|Z_i)} - \frac{Y_i(1- T_i)}{1-P(T_i=1|Z_i)}
\end{equation}

The final estimator, augmented-IPW (AIPW), interpolates between the two estimators \cite{robins1994estimation,van2011targeted}. 
It has been shown that the AIPW estimator satisfies double robustness: it retrieves 
consistent estimates if either the propensity score or outcome model is correct even if the other is misspecified.
It is:
\begin{equation}
\begin{split}
\aipw =  \frac{1}{n} \sum_{i=1}^{n} &\frac{T_i (Y_i - Q(Z_i, 1))}{P(T_i=1|Z_i)} - \frac{1 - T_i (Y_i - Q(Z_i, 0))}{1-P(T_i=1|Z_i)} \\
& + Q(Z_i, 1) - Q(Z_i, 0)
\end{split}
\end{equation}

All three estimators rely on measurements of confounders $Z$.
We will see that in debate threads, we must recover the confounders from high-dimensional text. 
A key idea of this paper is to use text data to find a representation for $Z$.
\section{Dataset}
\label{sec:data}

\begin{figure*}[h!]
	\centering
	\includegraphics[scale=0.6]{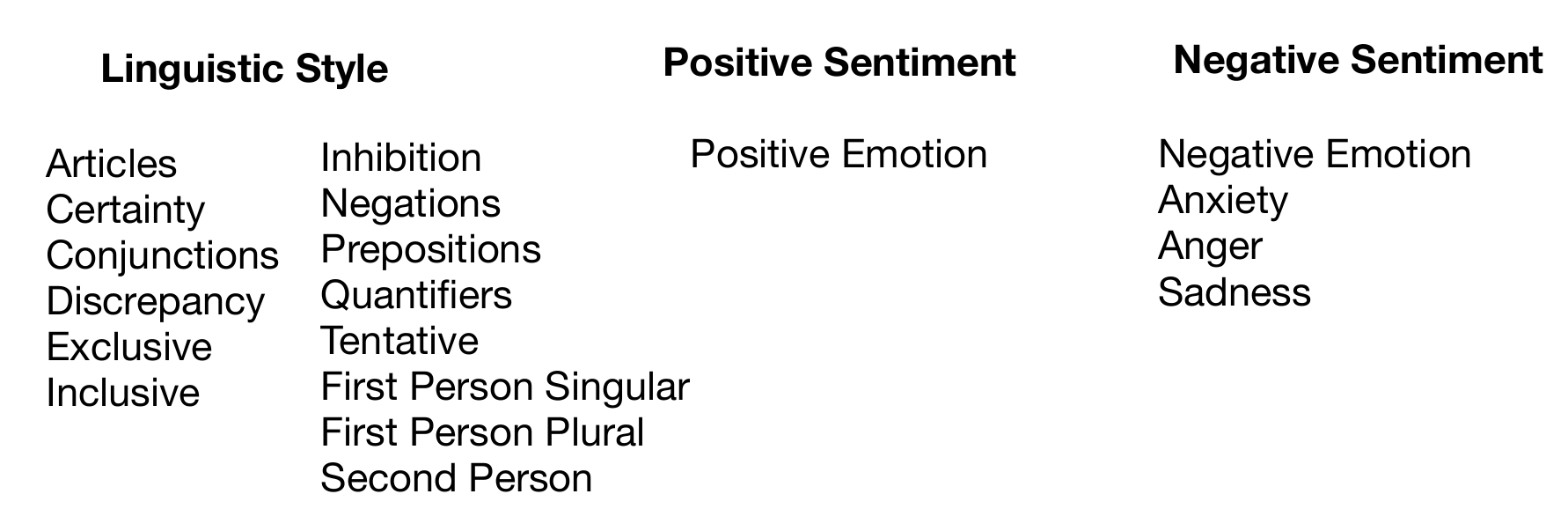}
	\caption{\label{fig:categories} We group LIWC categories into three types: linguistic style, positive sentiment, negative sentiment.
		We use the groupings for vector representations of posts which we can use to measure the outcomes of interest: 
		changes between the first and last post of a triple.}
\end{figure*}


To estimate the ATE of tone, we use the \Forumcom~corpus collected and annotated as part of the Internet Argument Corpus \cite{Walkeretal12c}.
\Forumcom~has been used to predict users' stances on topics, disagreements between users, and sarcasm use \cite{lukin2013really,Walkeretal12a,Walkeretal12b,sridhar2015joint}.

\Forumcom~is a collection of debate discussions, each belonging to a topic such as ``evolution'' or ``climate change.''
For some pairs of posts called \textbf{quote-response} pairs, \Forumcom~includes annotations about the reply, obtained using Amazon Mechnical Turk. 
A quote-response pair is a post and its reply where the replier quotes the original poster and responds directly to the quoted statement. 
The response is annotated by multiple annotators along four dimensions which we refer to as \textbf{reply types}: nasty/nice, attacking/reasonable, emotional/factual, questioning/asserting. 
Each reply type has two opposing polarities (e.g., nasty or nice) which we refer to as its \textbf{tone}. The annotation score for each type ranges from -5 to 5, where negative values correspond to the antagonistic tone such as nasty or attack and positive values map to tones such as reasonable or factual.

We select the four debate topics with the most quote-response annotations: ``abortion'', ``gay marriage'', ``evolution'', and ``gun control''. 
Each debate topic has on roughly 1200 quote-response annotations.
Following prior work, we use the mean score across annotators and discard annotations with a mean score between -1 and 1 \cite{MisraWalker13,lukin2013really,sridhar2015joint}.
In the next section, we formalize the use of these annotations as treatments to estimate causal effects.
\section{Problem Statement}
\label{sec:problem}

To study causal effects in debate threads, we first introduce \textbf{post triples}.
A \textbf{post triple} $t_i = (p^1_i, p^2_i, p^3_i)$ is an ordered sequence of three posts where each post $p^j_i$ belongs to the $i$-th triple and appears $j$-th in the sequence. 
The author of post $p^j_i$ is denoted by $a^j_i$.
The triples we consider have the property that $a^1_i = a^3_i$, i.e., the same user authors the first and last posts. 
Based on the discussion in which the triple appears, the triple $t_i$ has a debate topic $\tau_i$.
We will refer to $p^1_i$ as the original post and to $p^2_i$ as the reply post.

Each triple we study has a quote-response annotation for $p^2_i$ towards $p^1_i$. 
Given the reply type $\alpha$ of the annotations and its mean score, we binarize the values by considering those $\leq -1$ as 0 and $\geq 1$ as 1. 
Replies are thus converted to binary negative or positive tone, such as nasty or nice.
For each triple $t_i$ and reply type $\alpha$, the tone of reply post $p^2_i$ toward $p^1_i$ gives the treatment assignment $T_i = R^\alpha_i \in \{0, 1\}$ for the triple.

\paragraph{Outcomes.} 
The next problem is to quantify the outcome of interest: changes between $p^3_i$ and $p^1_i$ after receiving reply $p^2_i$.
We rely on the Linguistic Inquiry and Word Count (LIWC) tool \cite{pennebaker2007liwc2007}. 
LIWC is a dictionary which maps an extensive set of English words to categories that capture both lexical and semantic choices. 
Several text classification and statistical analysis tasks have represented posts with counts of LIWC categories \cite{anand2011cats,Abbottetal11,Walkeretal12a,danescu2011mark}.

We first combine LIWC categories into groups which we call \textbf{category types} that measure positive sentiment, negative sentiment, and linguistic style.
\figref{fig:categories} shows each category type.
For the sentiment groupings, we select the categories related to positive and negative emotion as listed on the LIWC website.
For linguistic style, we use the categories identified in prior work for linguistic style accommodation \cite{danescu2011mark}.

Given a category type, the frequency of words in $p^j_i$ belonging to each category gives a vector representation of the post.
We can construct such vector representations for $p^1_i$ and $p^3_i$.
Formally, for a category type $c$ and reply type $\alpha$, the outcome $Y^{c,\alpha}_i$ 
for triple $t_i$ is the Euclidean distance between the vector representations for $p^1_i$ and $p^3_i$.
This strategy suggests many possible vector representations of posts including word embeddings \cite{mikolov2013distributed}. 

We state the ATE estimatation problem for these debate triples. For all configurations of $c$ and $\alpha$, we estimate:
$$\psi = \mathbb{E} \big[\mathbb{E}[Y^{c,\alpha}|Z, R^\alpha=1] - \mathbb{E}[Y^{c,\alpha}|Z, R^\alpha=0]]$$
This estimates the mean difference in text changes between users receiving a positive-tone reply and those receiving negative-tone ones.
The main challenge is to find a representation for $Z$ that captures plausible confounders in debates.
\section{Constructing Confounder Representations}
\label{sec:approach}

In a post triple of interest, the debate topic, latent ideologies of each author within the topic, and sentiment of the original author are confounders.
That is, these variables plausibly influence both the treatment (reply tone) and the outcome (change between posts).
Prior work has shown that text in political debates can be mapped into a lower dimensional space
that corresponds to the moral or ideological facets of that debate topic 
\cite{johnson2018classification,misra2015using,iyyer2014political,boydstun2013identifying}. 
Unsupervised approaches have been used to discover word-clusters that correspond to these frames directly from text \cite{iyyer2014political}. 
Here, we fit an unsupervised generative model of text to learn ideology representations.

\paragraph{Ideology Representation.} 
We use the latent Dirichlet allocation (LDA) topic model \cite{blei2003latent} to recover unobserved ideologies.
The observations $w_{ij}$ are counts of word $j$ in document $i$.
The generative process is:
\begin{eqnarray}
\beta_{k} &\sim& \text{Dirichlet}(\gamma) \\
\theta_{i} &\sim& \text{Dirichlet}(\alpha) \\
z_{ij} &\sim& \text{Multinomial}(\theta_i )\\
w_{ij} &\sim& \text{Multinomial}(\beta_{z_{ij}})
\end{eqnarray}
Each of the $k$ latent topics $\beta_k$ is a distribution over the words in the vocabulary.
Each document-level latent variable $\theta_i$ is a distribution over topics.
For a document $i$, each word $w_{ij}$ is drawn by sampling a topic from the document's distribution over topics and then sampling a word from that topic.

The posterior expected values $\mathbb{E}[\theta_i]$ and $\mathbb{E}[\beta_j]$ converge to the optimal values of $\theta_i$ and $\beta_j$.
This convergence property allows us to substitute $\mathbb{E}[\theta_i]$ as a confounder for each post.
LDA is fit using variational inference.

We fit LDA for each debate topic $\tau$ with the observed word counts across posts from that topic.
By conditioning on $\tau$, the confounder representation incorporates both the debate topic and finer-grained ideology.
The inferred mean proportions over latent topics $\mathbb{E}[\theta_i]$ is the embedding for each post.
For each triple $t_i$, we concatenate the embeddings for posts $p^1_i$ and $p^2_i$ to include in the confounder $Z_i$.
Since the embeddings aim to approximate ideologies, including both $p^1_i$ and $p^2_i$ embeddings
helps to further deconfound the effect of users' opposing or similar views on tone and word change.

\paragraph{Sentiment Representation.}
To represent the sentiment encoded in $p^1_i$, we use the same LIWC category types as we do for the outcomes.
As before, we compute the frequency of words in $p^1_i$ that belong to each category.
This gives us a vector representation of sentiment which we include in $Z_i$.


\section{Empirical Results}
The difficulty in validating causal effects, particularly in debates, is that there is no ground truth. 
Typically, causal estimation procedures are validated using simulated data but for text,
realistic generative models do not exist. 
Thus, one of the paper’s contributions is developing an evaluation strategy for text-based causal inferences. 
Our approach is three-pronged: 1) we assess the predictive performance of the key ingredients for estimation, the propensity score and expected outcome models; 
2) we manually inspect the latent ideological topic ; 3) we compare causal effects across multiple estimators and against a naive confounder representation.

We found that: 1) causal estimation using our confounder representation reduces bias in the ATE estimates compared to using a naive confounder; 
2) if we had not compared multiple estimators and instead used a single estimator like the high-variance IPTW (a common practice), we would have incorrectly reported effects;
3) the estimates suggest that emotional/factual and questioning/asserting tones elicit changes in linguistic style and emotion while nice/nasty or reasonable/attacking tones show no effect.
Code and data to reproduce all results are available. \footnote{\texttt{github.com/dsridhar91/debate-causal-effects}}

\paragraph{Methods and Metrics.}
Using the latent confounder representation proposed, the goal is to estimate the ATE $\psi$ of reply tone $R_\alpha$ on outcome $Y^{c,\alpha}$. 
In the empirical results below, we estimate $\psi$ using three estimators ($\mle, \ipw, \aipw$) for all configurations of LIWC category type $c$ and $\alpha$ that yield different treatments and outcomes. 
We report the unadjusted estimate, $\big[\mathbb{E}[Y|T=1] - \mathbb{E}[Y|T=0]$,
which will be biased.
We compare against a naive representation, $Z$ - Debate Topics Only, which only uses the debate topic $\tau_i$ without
finer-grained ideologies.

We fit the propensity score $P(T=1|Z)$ (used by $\ipw, \aipw$)
with logistic regression using the observed treatments and constructed confounder representations.
We fit the expected outcomes, $Q(Z, T=0), Q(Z, T=1)$ (used by $\mle, \aipw$) with linear regression.

\paragraph{Experimental Setup.}
Besides the processing of quote-response pair annotations described in the Data section,
we prepare the posts to fit LDA. We obtain all unigram tokens
after lemmatizing and removing stop words from posts across all discussions for a given topic.
We retain only those tokens which occur in more than 2\% but in fewer than 80\% of posts. 
For each topic, this yields a document-term-frequency matrix of roughly 30, 000 posts
and 400 remaining terms after pre-processing.
We fit LDA with $k=50$ topics.
For each reply type, the ATE is averaged over roughly 1500 triples.
For the AIPW estimator $\aipw$, we use a variant proposed to improve finite sample performance \cite{van2011targeted}.
\begin{table*}[h!]
	\begin{center}
		\begin{tabular}{c c c c c c c c}
			\multicolumn{8}{c}{Performance of Outcome (Pos. Sent, RMSE) and Propensity Models (F1)}\\
			\toprule 
			\textbf{Reply type} & \multicolumn{3}{c}{$Z$ - Debate Topics Only} && \multicolumn{3}{c}{$Z$ - Full} \\
			\midrule
			& \textbf{$Q(Z,T=1)$} &  \textbf{$Q(Z,T=0)$} & \textbf{$P(T=1|Z)$} && \textbf{$Q(Z,T=1)$} & \textbf{$Q(Z,T=0)$ }& \textbf{$P(T=1|Z)$} \\
			\midrule
			Nasty/Nice & 3.5 & 2.7 & 0.89 && 2.6  & 2.4 & 0.89 \\
			Attacking/Reasonable & 3.6 & 2.7 & 0.81  && 3.0 & 2.7 & 0.81 \\
			Emotional/Factual & 2.4 & 5.1 & 0.69  &&  2.2 & 5.1 & 0.72 \\
			Questioning/Asserting & 3.1 & 4.4 & 0.80 && 2.9 & 4.1 & 0.79\\
			\bottomrule
		\end{tabular}
	\end{center}
	\caption{\label{tab:crossval} The expected outcome models (shown here for positive sentiment) using $Z$-Full generally improves over using $Z$-Debate Topic Only. The propensity score model performs comparably in both cases. We evaluate the models using 5-fold cross-validation. For expected outcomes, we report RMSE and F1 for the propensity score.}
 \end{table*}

\begin{figure*}
	\centering
	\includegraphics[scale=0.5]{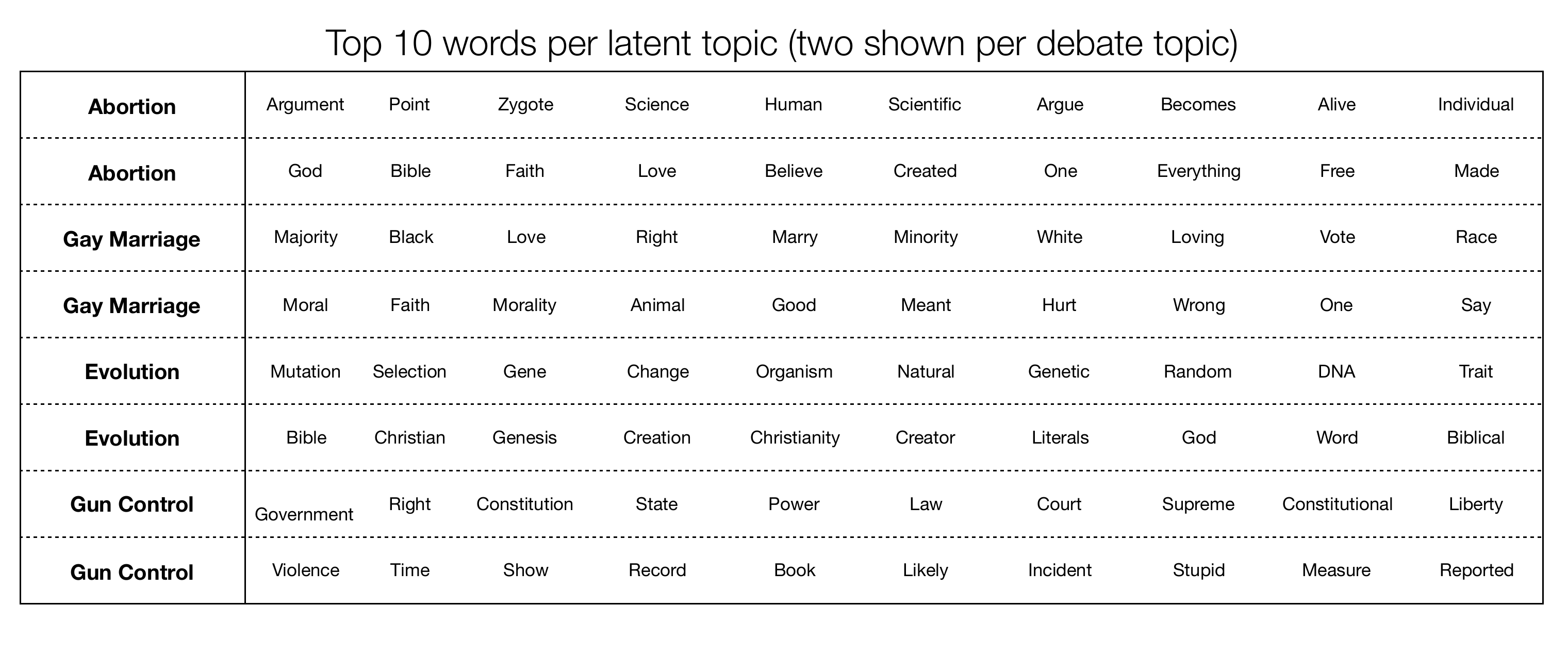}
	\caption{\label{fig:topics} Top words across latent topics from each debate
	topic chosen to illustrate ideologies captured by LDA. The latent topics are suggestive of viewpoints like
	morality and faith versus science and evidence.}
\end{figure*}

\paragraph{Performance of Outcome and Propensity Models.}
The first step to validating ATEs is to verify that the models used in various estimators fit the observations well.
\tabref{tab:crossval} gives the root mean squared error (RMSE) and F1 for the expected outcomes $Q(Z,T)$ and propensity scores $P(T=1|Z)$, respectively.
We perform five-fold cross-validation on the triples. 
We report performances for all reply types $\alpha$ but for the outcome model which depends also on the LIWC category type, 
we show scores for positive sentiment outcomes for conciseness. 
Our code will reproduce the other RMSEs.

For the positive sentiment expected outcomes, our confounder $Z$-Full predicts with lower RMSE than $Z$-Debate Topics Only.
The F1 score is also slightly improved by using $Z$-Full over $Z$-Debate Topics Only when predicting emotional/factual.
While this is reassuring, for causal inference, unbiased estimation is more important than predictive
performance.
Below, we investigate causal estimation, where $Z$-Debate Topics Only shows undesirable consequences.

\begin{table*}[h!]
	\footnotesize
	\begin{center}
		\begin{tabular}{c c c c c c c c}
			\multicolumn{8}{c}{\textit{ATE (and Standard Error) for Nasty/Nice}}\\
			\toprule
			\textbf{Estimator} & \multicolumn{3}{c}{$Z$ - Debate Topics Only} && \multicolumn{3}{c}{$Z$ - Full}\\ \\
			\midrule
			& \textbf{Pos.} & \textbf{Neg.} & \textbf{Ling.} &&  \textbf{Pos.} & \textbf{Neg.} & \textbf{Ling.}\\
			\midrule
			Unadjusted & 0.0 & -0.8 & -0.6 && 0.0 & -0.8 & -0.6 \\
			$\mle$ & 0.0 (0.0) &  -0.8 (0.1) & -0.6 (0.1) && 0.0 (0.1) & -0.3 (0.1) & -0.3 (0.1)  \\
			$\ipw$ & 0.0 (0.2)  & -0.7 (0.35)  & -0.6 (1.0) && 0.0 (0.3) & -0.2 (0.3)  & -0.1 (1.0) \\
			$\aipw$ & 0.0 (0.2) & -0.8 (0.3)  & -0.6 (0.5)  && 0.0 (0.1) &  -0.3 (0.2)  & -0.3 (0.4) \\
			\bottomrule
		\end{tabular}
	\end{center}
	\caption{\label{tab:ateNN} The debate topics-only approach can overestimate treatment effects; it remains more biased (compared to the unadjusted estimate) than using $Z$-Full. We report ATE (and standard error) for Nasty/Nice reply type.}
\end{table*}

\begin{table*}[h!]
	\begin{center}
		\small
		\begin{tabular}{c c c c c c c c c c c c}
			\multicolumn{12}{c}{\textit{ATE (and Standard Error) for Remaining Reply Types}}\\
			\toprule
			\textbf{Estimator} & \multicolumn{3}{c}{\textbf{Attacking/Reasonable}} && \multicolumn{3}{c}{\textbf{Emotional/Factual}} && \multicolumn{3}{c}{\textbf{Questioning/Asserting}} \\
			\midrule
			& \textbf{Pos.} & \textbf{Neg.} & \textbf{Ling.} &&  \textbf{Pos.} & \textbf{Neg.} & \textbf{Ling.} &&  \textbf{Pos.} & \textbf{Neg.} & \textbf{Ling.}\\
			\midrule
			Unadjusted & -0.1  & -0.6 & -0.6 && -1.0 & -0.8 & -2.3 && -0.5 & -0.2 & -1.7 \\
			$\mle$ & 0.1 (0.1) & -0.2 (0.1) & -0.2 (0.1) && {\bf -0.6 (0.1)} & {\bf -0.3 (0.1)} & {\bf -1.4 (0.1)} && {\bf -0.3 (0.1) }& -0.2 (0.1) & {\bf -1.2 (0.1)} \\
			$\ipw$ & 0.1 (0.2) & -0.1 (0.3) & -0.4 (0.4) && -0.4 (0.3) & -0.2 (0.2) &  -0.7 (0.8)  && -0.3 (0.3)  & -0.2 (0.2)  & -1.1 (0.8) \\ 
			$\aipw$ & -0.1 (0.1) & -0.2 (0.9)  & -0.4 (0.4) &&  {\bf -0.6 (0.2)} & {\bf -0.3 (0.1)} & {\bf -1.2 (0.3)} && -0.3 (0.2) & -0.2 (0.1) & {\bf -1.2 (0.3)}\\
			\bottomrule
		\end{tabular}
	\end{center}
	\caption{\label{tab:ateAll} Factual and asserting tones result in the first dialogue participant significantly decreasing changes in linguistic style. Factual tones may provoke
		decreased change in sentiment. We report ATE (and standard error) for all remaining reply types. Bolded numbers indicate that the ATE is significantly greater than zero.}
\end{table*}

\paragraph{Latent Ideologies.}

Even if the ideology representations inferred using LDA are useful for predictive performance above, we carefully
inspect the latent topics found by LDA.
Since we assumed that ideology is a confounder, we want the latent topics to approximate ideologies.
We inspected the top ten words associated with each latent topic across all debate topics.
\figref{fig:topics} shows these words for two latent topics from every debate topic as illustrative examples
of the ideological views found.
For example, in gun control debates, LDA finds topics that align with constitutional rights to bear arms
and in evolution debates, there are contrasting topics that align with creationist versus scientific views.
Our code includes simple visualization to inspect all latent topics.
\paragraph{ATE Estimation.}
We use $Z$-Debate Topic Only and $Z$-Full and apply the three estimators $\mle, \ipw$ and $\aipw$. We compare
these estimates against the unadjusted estimate.
\tabref{tab:ateNN} shows the ATEs (and standard error) for the nasty/nice reply type.
When confounders are missing from adjustment, we expect the estimate to be closer to the biased, unadjusted estimate.
Indeed, the results show that using $Z$ - Debate Topics Only, omitting sentiment and ideology, consistently
yields estimates which are closer to the unadjusted estimate than using $Z$ - Full.
This is a key finding: estimation bias is reduced with the finer-grained confounder representation.
After adjusting for $Z$ - Full, the effects on dialogue outcomes from nasty/nice tones are not significant.

In \tabref{tab:ateAll}, we proceed with our confounder representation $Z$ - Full and study the remaining reply types: attacking/reasonable, emotional/factual, questioning/asserting.
The $\mle$ and $\aipw$ estimators find significant effects, particularly for factual and authoritative tones.
The $\ipw$ estimator yields similar ATE estimates but suffers from high variance.
This is another key finding that comparing 
multiple estimators provides a form of validation: the propensity score-based $\ipw$ estimator is known to have high variance, and without the two remaining estimators, we may have concluded that no significant effects occur.

The $\mle$ and $\aipw$ estimators suggest that factual and asserting tones cause decreased changes in linguistic style: 
users' second posts remain closer to their original posts on average across triples. 
Further, these estimators suggest that both positive and negative sentiment changes are decreased when the tone is factual.
The results suggest that users change their overall sentiment
less when responding to factual arguments instead of emotionally charged ones.
However, users may also maintain their original linguistic styles more when responding to factual or asserting tones.
This finding on the role of factual and asserting tones may point to in-depth followup studies on persuasion and argumentation in debates.

Finally, the empirical studies reveal unexpected findings about causal estimation of treatments effects in debates.
Interestingly, the choice of outcome representation matters: \tabref{tab:ateAll} in particular shows that changes to linguistic style are affected more than sentiment changes. 
A single outcome representation which had concatenated all LIWC categories or used word embeddings may have yielded different results.
\section{Discussion}
\label{sec:conclusion}

We study treatment effects in debates by estimating unobserved confounders from sequences of posts. 
We examine these interpretable embeddings in debates to find that they match known ideological views.
The exercise of estimating treatment effects yields results of two flavors: 1) evidence that factual replies cause decreased change in linguistic style and sentiment, and 2) guidelines for practioners to estimate treatment effects from social media text.

We highlight areas of future study. 
In this work, we focus on ideology and sentiment as confounders. 
It is interesting to consider possible confounding from
posts' timing and position in a discussion thread.
A fruitful area of research is learning deep confounder (and outcome) representations for text
while maintaining model interpretability, which we show in this paper is important for validating findings.
Finally, validating causal effects in questions of social science research remains an open problem.
Simulating outcomes from text is a line of future work.
\section*{Acknowledgements}
This work is supported by NSF grants CCF-1740850 and IIS-1703331.

\bibliographystyle{named}
\bibliography{sridhar-ijcai19}
\end{document}